\documentclass[runningheads]{llncs}

\usepackage{svg}
\usepackage[fleqn]{amsmath}
\usepackage{libertine}
\usepackage{epsfig}
\usepackage{graphicx}
\usepackage{amsmath}
\usepackage{amssymb}
\usepackage{enumitem}
\usepackage{mathtools}

\DeclareMathOperator*{\argmax}{arg\,max}

\usepackage{footmisc}

\usepackage{mathrsfs}
\usepackage{adjustbox}
\usepackage{tabularx}
\usepackage{booktabs}
\usepackage{multirow}
\usepackage{flushend}

\newcolumntype{m}[1]{>{\centering\arraybackslash}p{#1}}

\definecolor{OliveGreen}{rgb}{0,0.6,0}
\definecolor{SoftRed}{rgb}{1,0.2,0.2}
\usepackage[pagebackref=true,breaklinks=true,letterpaper=true,colorlinks,bookmarks=false,citecolor=OliveGreen,linkcolor=SoftRed]{hyperref}

\usepackage{xcolor}

\usepackage{color}
\usepackage[width=122mm,left=12mm,paperwidth=146mm,height=193mm,top=12mm,paperheight=217mm]{geometry}

\begin{document}
\pagestyle{headings}
\mainmatter

\title{OPAL-Net: A Generative Model for Part-based Object Layout Generation}

\titlerunning{OPAL-Net}

\author{Rishabh Baghel \and
Ravi Kiran Sarvadevabhatla }
\authorrunning{Baghel and Sarvadevabhatla}
\institute{Centre For Visual Information Technology\\IIIT Hyderabad, INDIA 500032 \\\email{baghelrishabha@gmail.com,ravi.kiran@iiit.ac.in}\\
\url{https://github.com/atmacvit/opalnet}}

\maketitle

\begin{abstract}
We propose OPAL-Net, a novel hierarchical architecture for part-based layout generation of objects from multiple categories using a single unified model. We adopt a coarse-to-fine strategy involving semantically conditioned autoregressive generation of bounding box layouts and pixel-level part layouts for objects. We use Graph Convolutional Networks, Deep Recurrent Networks along with custom-designed Conditional Variational Autoencoders to enable flexible, diverse and category-aware generation of object layouts. We train OPAL-Net on PASCAL-Parts dataset. The generated samples and corresponding evaluation scores demonstrate the versatility of OPAL-Net compared to ablative variants and baselines.
\keywords{generative model, part-based object representation, GCN, VAE, semantic layout generation}
\end{abstract}

\section{Introduction}

Many recent and exciting successes for generative models have been associated with generation of realistic images using top-down guidance from text~\cite{hong2018inferring}, scene attributes~\cite{ritchie2019fast} or pixel-level semantic conditioning~\cite{park2019SPADE}. Generative models specifically for objects have been relatively less explored. Developing such models is very challenging since the compositional primitives (parts) for objects need to obey stricter shape and meronymic\footnote{Meronym is a linguistic term for expressing part-to-whole relationships.} constraints compared to scenes. For an \texttt{airplane}, the tail needs to be attached to the fuselage, wings need to be equally shaped and located symmetrically relative to the fuselage. For a \texttt{cow}, eyes need to be inside the head and horns need to be attached at the top. In these terms, even a minor amount of misalignment in the generated object is immediately obvious to human eye due to the Law of Pragnanz.

Generative models do exist for objects with aligned part configurations (e.g. faces~\cite{bessinger2019generative,he2019attgan}) or with an accompanying text description (e.g. birds~\cite{yin2019semantics,zhang2018stackgan++}). However, part-based generative models for general collection of highly articulated objects in 2-D have been relatively less explored. Designing such models, with semantic part-level information guiding the object generation process, can lead to increased diversity and appearance quality in the generated samples. Inspired by the success of scene generation approaches driven by top-down guidance~\cite{hong2018inferring,park2019SPADE}, we take a step towards generative models for objects by developing a novel hierarchical, class-aware generative model for object layouts. Starting from a specified object category and an associated part list, we employ a first-level generative model which stochastically generates a list of part bounding boxes (Sec. \ref{sec:boxvae}). The category, part list and generated bounding boxes are used to condition a second-level model which stochastically generates semantic part maps for the specified object class (Sec.~\ref{sec:labelmapvae}). Our unified approach enables layout generation for \textit{any} of the object classes from a single model instead of maintaining per-class models~\cite{zhang2018stackgan++,li2019layoutgan}. Additionally, our hierarchical design enables diverse and interactive generations via its ability to incorporate a user-specified set of parts.

 Overall, our model paves the way for enabling compact, hierarchically configurable and truly end-to-end models of scene generation. In such a scenario, the generation process can be controlled at class and part level for objects and subsequently conditioned on objects at the overall scene level.

\section{Related Work}

\noindent \textbf{Layout Generation:} The approach of Li et al.~\cite{li2019layoutgan} involves initial production of randomly placed graphic elements whose attributes are refined via self-attention to generate document layouts. A number of works study stochastic layout generation for scenes. Jyothi et al.~\cite{Jyothi2019ICCV} provide a good summary of representative works on the topic. As part of their overall strategy in generating 2-D scenes, Hong et al.~\cite{hong2018inferring} and Jyothi et al.~\cite{Jyothi2019ICCV} demonstrate the benefit of stochastically modelling the distribution of structural elements (object bounding boxes) and their attributes (class labels and instance counts). We too employ the approach of intermediate stochastic structure generations but with larger number of levels in the generative hierarchy and with object parts as structural elements.

\noindent \textbf{2-D object generative models:} Within the 2-D realm, approaches are designed for a specific object type (e.g. faces~\cite{bessinger2019generative}, birds~\cite{yin2019semantics,zhang2018stackgan++}, flowers~\cite{park2018mc}) and usually involve text or pixel-level conditioning~\cite{isola2017image}. In some cases, attributes (including parts) are used to generate these specific object types~\cite{he2019attgan,yan2016attribute2image}.  Apart from these, objects are usually generated as intermediate component structures by 2-D scene generation approaches~\cite{hong2018inferring,park2019SPADE,Jyothi2019ICCV}. Since objects usually occupy a smaller spatial extent relative to the scene dimensions, the resulting generations tend to be blurry and deficient in variety. Part-aware models for general 2-D object collections do not exist, to the best of our knowledge.

\noindent \textbf{3-D object generative models:} A variety of interesting part-based generative approaches exist for 3-D objects~\cite{SAGnet19,Mo2019CVPR,mo2019structurenet,li2017grass,nash2017shape}. Unlike our unified model, these approaches train a separate model for each object. Also, the number of object categories, maximum number of parts and variation in intra-category spatial articulation in these approaches is generally smaller compared to our setting.

\noindent \textbf{Graph Convolutional Networks (GCNs):} GCNs have recently emerged as a popular framework for working with graph-structured data and predominantly for discriminative tasks~\cite{zhangdual,verma2018feastnet,yang2019auto,zhao2019semantic}. In a generative setting, GCNs have been applied for scene graph generation~\cite{yang2018graph,gu2019scene}. In our work, we use GCNs which operate on part-graph object representations. These representations, in turn, are used to train a Variational Auto Encoder (VAE) generative model. GCN-VAE have been previously used for modelling citation networks~\cite{kipf2016variational,yang2018meta}, gene interaction networks~\cite{yang2018meta} and molecular design~\cite{liu2018constrained}. To the best of our knowledge, we are the first to introduce a conditional variant of GCN-VAE for images and specifically, for 2-D layout generation.

\section{Overview}
\label{sec:overview}

We begin with a brief overview of Variational Auto Encoder (VAE)~\cite{kingma2013auto} and its extension Conditional Variational Auto Encoder (CVAE)~\cite{sohn2015learning} which we employ as our base generative model. Subsequently, we provide a summary of our overall generative pipeline (Sec. \ref{sec:approachsummary}).

\subsection{VAE and CVAE}
\label{sec:vae}

\noindent \textbf{VAE:} Let $p(\mathsf{d})$ represent the probability distribution of data. VAEs transform the problem of generating samples from $p(\mathsf{d})$ into that of generating samples from the likelihood distribution $p_{\theta}(\mathsf{d}|z)$ where $z$ is a low-dimensional latent surrogate of $\mathsf{d}$, in turn modelled as a standard normal distribution, i.e. $p(z) = \mathcal{N}(0,I)$. Now, obtaining an accurate latent representation for a given data sample $\mathsf{d}$ is possible if we have the exact posterior distribution  $p(z|\mathsf{d})$. Since the latter happens to be intractable, a variational approximation $q_{\phi}(z|\mathsf{d})$, which is easier to sample from, is used. Typically, the distribution is modelled using an `encoder' neural network parameterized by $\phi$. Similarly, the likelihood distribution is modelled by a `decoder' neural network parameterized by $\theta$. To jointly optimize for $\theta$ and $\phi$, the so-called evidence lower bound (ELBO) of the data's log distribution ($\text{log } p(\mathsf{d})$) is sought to be maximized. The ELBO is given by $\mathcal{L}(\mathsf{d};\phi,\theta) = \mathbb{E}_{q_{\phi}(z|\mathsf{d})} [\text{log } p_{\theta}(\mathsf{d}|z)] - \lambda KL(q_{\phi}(z|\mathsf{d})\lvert \lvert p(z))$ where KL stands for KL-divergence and $\lambda >0$ is a tradeoff hyperparameter.

\noindent \textbf{CVAE:} Conditional VAEs are an extension of VAEs which can incorporate auxiliary information $\mathsf{a}$ as part of the encoding and generation process. As a result, the decoder now models $p_{\theta}(\mathsf{d}|z,\mathsf{a})$ while the encoder models $q_{\phi}(z|\mathsf{d},\mathsf{a})$. Accordingly, ELBO is modified as:

\begin{align}
\mathcal{L}(\mathsf{d,a};\phi,\theta) = \mathbb{E}_{q_{\phi}(z|\mathsf{d,a})} [\text{log } p_{\theta}(\mathsf{d}|z,\mathsf{a})] - KL(q_{\phi}(z|\mathsf{d,a})\lvert \lvert p(z|\mathsf{a}))
\end{align}

\noindent For our problem setting, the conditioning is performed as a gating operation where auxiliary information $\mathsf{a}$ either modulates a feature representation of input (encoder phase) or modulates the latent variable $z$ (decoder phase).

We modify CVAE and design an even more general variant where the conditioning used for encoder and decoder can be different. For additional details on VAE and CVAE, refer to the excellent tutorial by Doersch~\cite{doersch2016tutorial}.

\begin{figure*}[!tbp]
  \centering
  \includegraphics[width=\textwidth]{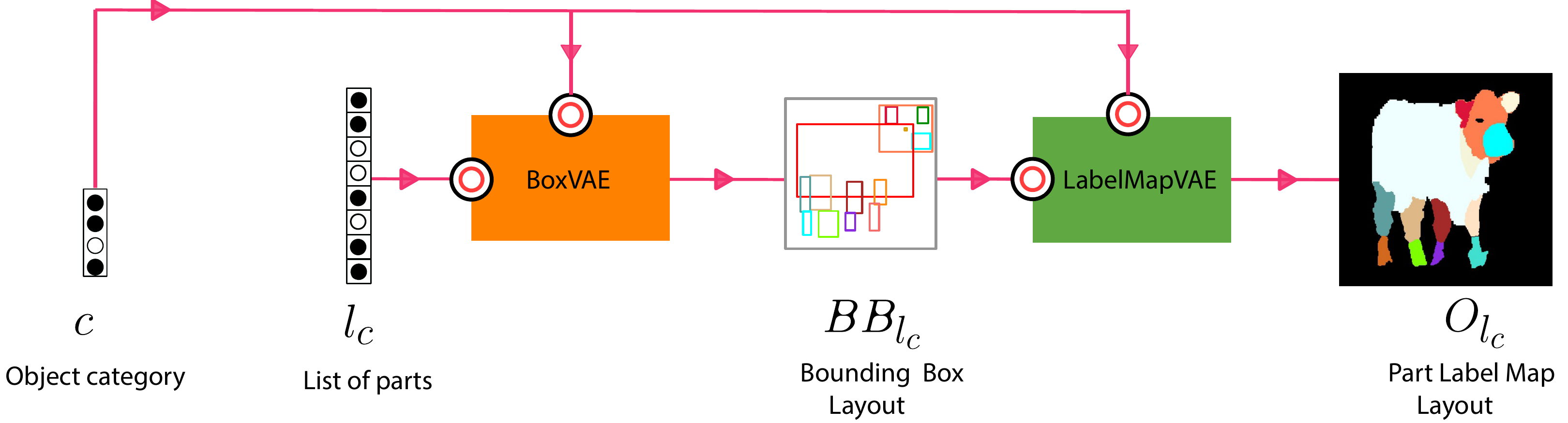}
  \caption{An illustration of the proposed hierarchical approach for object layout generation. Given an object label $c$ and a list of plausible object parts $l_c$, a bounding box layout of the object ${BB}_{l_c}$ is stochastically generated by BoxVAE (Sec. \ref{sec:boxvae}). The part-labelled bounding box layout is used by LabelMapVAE module (Sec. \ref{sec:labelmapvae}) to guide the stochastic generation of the final object layout $O_{l_c}$. The black-red circles indicate conditioning using object attributes during generation.}
  \label{fig:overview}
\end{figure*}

\subsection{Summary of our approach}
\label{sec:approachsummary}

Suppose we have $M$ object categories and we wish to generate an object layout for category $c,1 \leqslant c \leqslant M$. Suppose  each object category is associated with a part list $L_c$. The category information ($c$) and a list of parts $l_c \subseteq L_c$ is used to condition BoxVAE, the first-level generative model (Sec. \ref{sec:boxvae}, orange box in Fig. \ref{fig:overview}) which stochastically generates part-labelled bounding boxes $\mathcal{BB}_{l_c}$. The generated part-labeled box layout $\mathcal{BB}_{l_c}$ and object label $c$ are used to condition the second level generative model called LabelMapVAE (Sec. \ref{sec:labelmapvae}, green box in Fig. \ref{fig:overview}) which generates the final class-aware per-pixel part-map layout $O_{l_c}$. In the section that follows, we provide architectural details for various generative modules that constitute OPAL-Net.

\section{OPAL-Net}
\label{sec:OPAL-Net}

\subsection{BoxVAE}
\label{sec:boxvae}

\begin{figure*}[!ht]
  \centering
  \includegraphics[width=\textwidth]{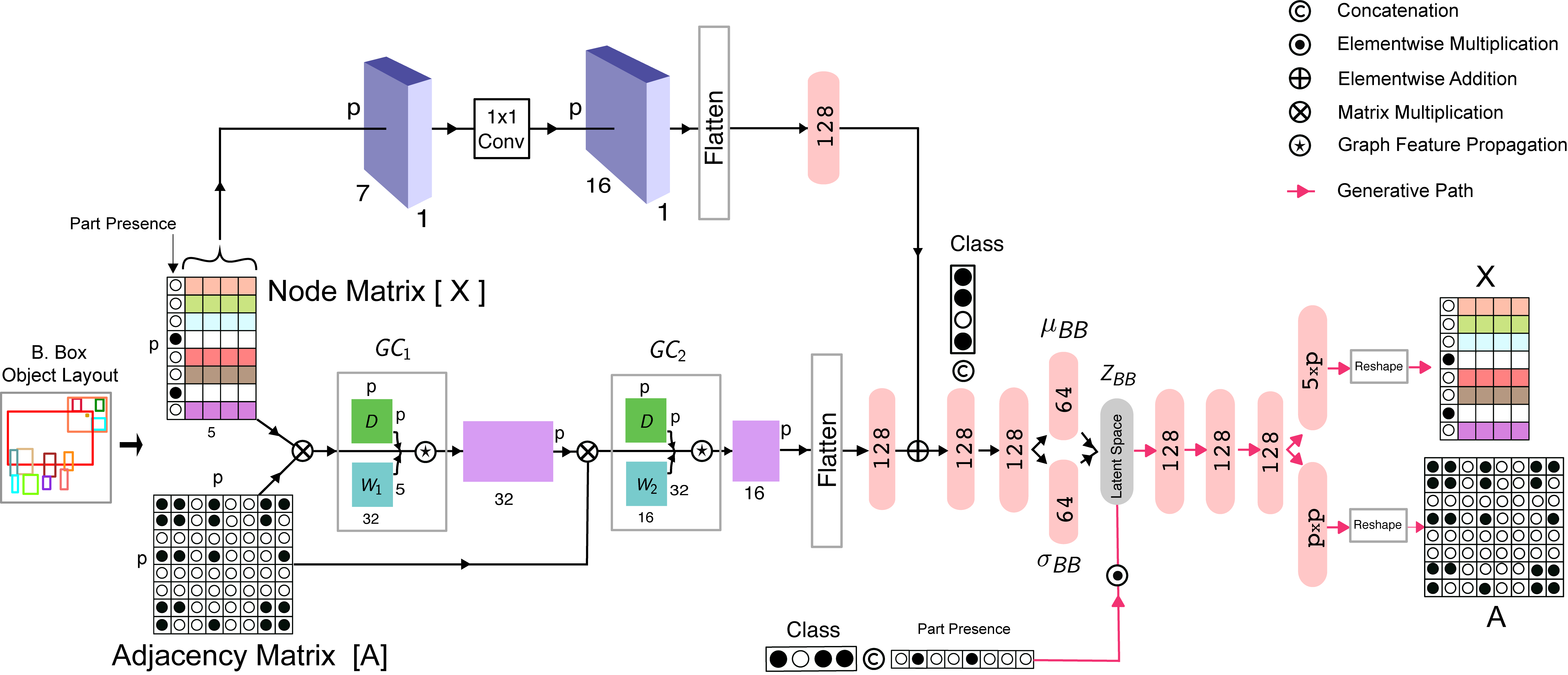}
  \caption{The architecture for BoxVAE (Sec. \ref{sec:boxvae}) which is trained to generate a graph representation of the object's bounding box layout. The pink arrows indicate the data flow for the generative model.}
  \label{fig:opalbox}
\end{figure*}

\noindent \textbf{Representing the bounding box object layout:} We design BoxVAE (Fig. \ref{fig:opalbox}) as a Conditional VAE which learns to stochastically generate part-labelled bounding boxes. To ensure realistic layouts, it is important to capture the semantic and structural relationships between object parts in a comprehensive manner. To meet this requirement, we model the bounding box object layout as an undirected graph. We represent the graph in terms of two matrices - Feature Matrix $X$ and Adjacency Matrix $A$~\cite{kipf2016variational,zhang2018graph}. Let $\mathsf{p}$ be the maximum possible number of parts across all object categories, i.e. $\mathsf{p} = \argmax_{c} l_c, 1 \leqslant c \leqslant M$. $X$ is a $\mathsf{p} \times 5$ matrix where each row corresponds to a part. For a given row $r$, a binary value $p_r \in \{0,1\}$ is used to record the presence or absence of the part in the first column. The next $4$ columns represent bounding box coordinates. For categories with part count less than $\mathsf{p}$ and for absent parts, the rows are filled with $0$s. For a given object category, the row indices consistently correspond with an arbitrarily ordered list of parts. The $\mathsf{p} \times \mathsf{p}$ binary matrix $A$ encodes the connectivity relationships between the object parts. Thus, we obtain the object part bounding box representation $\mathbb{G} = (X,A)$.

\noindent \textbf{Graph Convolutional Network (GCN):} A GCN takes a graph of the form $\mathbb{G}$ as input and computes hierarchical feature representations at each node in the graph while retaining the original connectivity structure. The feature representation at the ($i+1$)-th layer of the GCN is defined as  $H_{i+1} = f(H_i, A)$ where $H_i$ is a $\mathsf{p} \times F_i$  matrix whose $j$-th row contains the feature representation for node indexed by $j$ ($1 \leqslant j \leqslant \mathsf{p}$). $f$ represents the so-called propagation rule which determines the manner in which node features of the previous layer are aggregated to obtain the current layer's feature representation. We use the following propagation rule~\cite{kipf2017semi}:

\begin{align}
    f(H_i,A) = \sigma({D}^{\frac{-1}{2}} \widetilde{A} {D}^{\frac{-1}{2}} H_i W_i)
\end{align}

\noindent where $\widetilde{A} = A + I$ represents the adjacency matrix modified to include self-loops, ${D}$ is a diagonal node-degree matrix (i.e. ${D}_{jj} = \sum_k  \widetilde{A}_{jk}$) and $W_i$ are the trainable weights for $i$-th layer. $\sigma$ represents a non-linear activation function (ReLU in our case). Also, $H_0 = X$ (input feature matrix). We use a two-layer GCN (refer to $GC_1,GC_2$ in Fig. \ref{fig:opalbox}).

\noindent \textbf{Encoding the graph feature representation:} The feature representation of graph $\mathbb{G}$ obtained from GCN is then mapped to the parameters of a $128$-dimensional diagonal Gaussian distribution ($\mu_{BB}(\mathbb{G}),\sigma_{BB}(\mathbb{G})$), i.e. the approximate posterior. This mapping is guided via class-level conditioning on a feature representation preceding the final encoder layer. In addition, the mapping is also conditioned using skip connection features. These skip features are derived by applying $1 \times 1$ convolutions along the spatial dimension of bounding box sub-matrix of input $\mathbb{G}$ (see top part of Fig. \ref{fig:opalbox}). In addition to providing part-focused guidance for layout encoding, the skip-connection also  helps avoid the issue of imploding gradients.

\noindent \textbf{Reconstruction:} The sampled latent variable $z$, conditioned using category and part presence variables $c,l_c$ is mapped by the decoder to the components of $G = (X,A)$. Recalling that the first column of $X$ is the binary part-presence vector, let us write $X = \left[ X_1 \lvert X_{bb}\right]$. $X_1$ is modeled as a factored multivariate Bernoulli distribution, i.e. $p_{{\tiny \theta_{b}}}(X_1|z,c,l_c) =  \prod_{k=1}^{\mathsf{p}} {D_{l_c^k}}^{l_c^k} {(1 - D_{l_c^k})}^{1-l_c^k}$ where $D_{l_c}$ is the corresponding output of the decoder.

To obtain accurate localization of part bounding boxes, we use two per-box instance-level losses: mean squared error $L_i^{MSE} = \sum_{j=1}^4 (X_{bb}^i[j] - \hat{X}_{bb}^i[j])^2$ and Intersection-over-Union $L_i^{IoU} = -ln (IoU(X_{bb}^i,\hat{X}_{bb}^i))$~\cite{UnitBox2016} between the predicted ($\hat{X}_{bb}$) and ground-truth ($X_{bb}$) bounding boxes. To impose additional structural constraints, we also use a pairwise MSE loss defined over centers of bounding box pairs. Denoting the Euclidean distance between centers of $m$-th and $n$-th bounding boxes as $d_{m,n}$, the pairwise loss is defined as $L_{m,n}^{MSE-c} = (d_{m,n} - \hat{d}_{m,n})^2$.

For the adjacency matrix ($A$), we use binary cross-entropy $L_{m,n}^{BCE}$ as the per-element loss. The overall reconstruction loss for the BoxVAE decoder for a given object can be written as:

\begin{multline}
     L_{\mathsf{rec}}^{\mathsf{BoxVAE}} = \\
     \underbrace{\frac{- ln (p_{{\tiny \theta_{b}}}(X_1|z,c,l_c))}{\mathsf{p}}}_{X_1} +
     \underbrace{\frac{\sum_{i=1}^{\mathsf{p}} ( L_i^{MSE} + L_i^{IoU} )}{\mathsf{p}} +   \frac{\sum_{m=1}^{\mathsf{p}} \sum_{\substack{n=1\\n \neq m}}^{\mathsf{p}} L_{m,n}^{MSE-c}}{\mathsf{p} (\mathsf{p}-1)}}_{X_{bb}} \\+
    \underbrace{\frac{\sum_{m=1}^{\mathsf{p}} \sum_{n=1}^{\mathsf{p}} L_{m,n}^{BCE}}{\mathsf{p}^2}}_{A}
\label{eqn:boxvae-recons-loss}
\end{multline}

Note that the decoder architecture is considerably simpler compared to encoder. As our experimental results shall demonstrate (Sec. \ref{sec:experiments}), the conditioning induced by category and part-presence, combined with the connectivity encoded in the latent representation $z$, turn out to be adequate for generating the object bounding box layouts despite the absence of graph unpooling layers in the decoder.

\subsection{LabelMapVAE}
\label{sec:labelmapvae}

\begin{figure*}[!t]
  \centering
  \includegraphics[width=\textwidth]{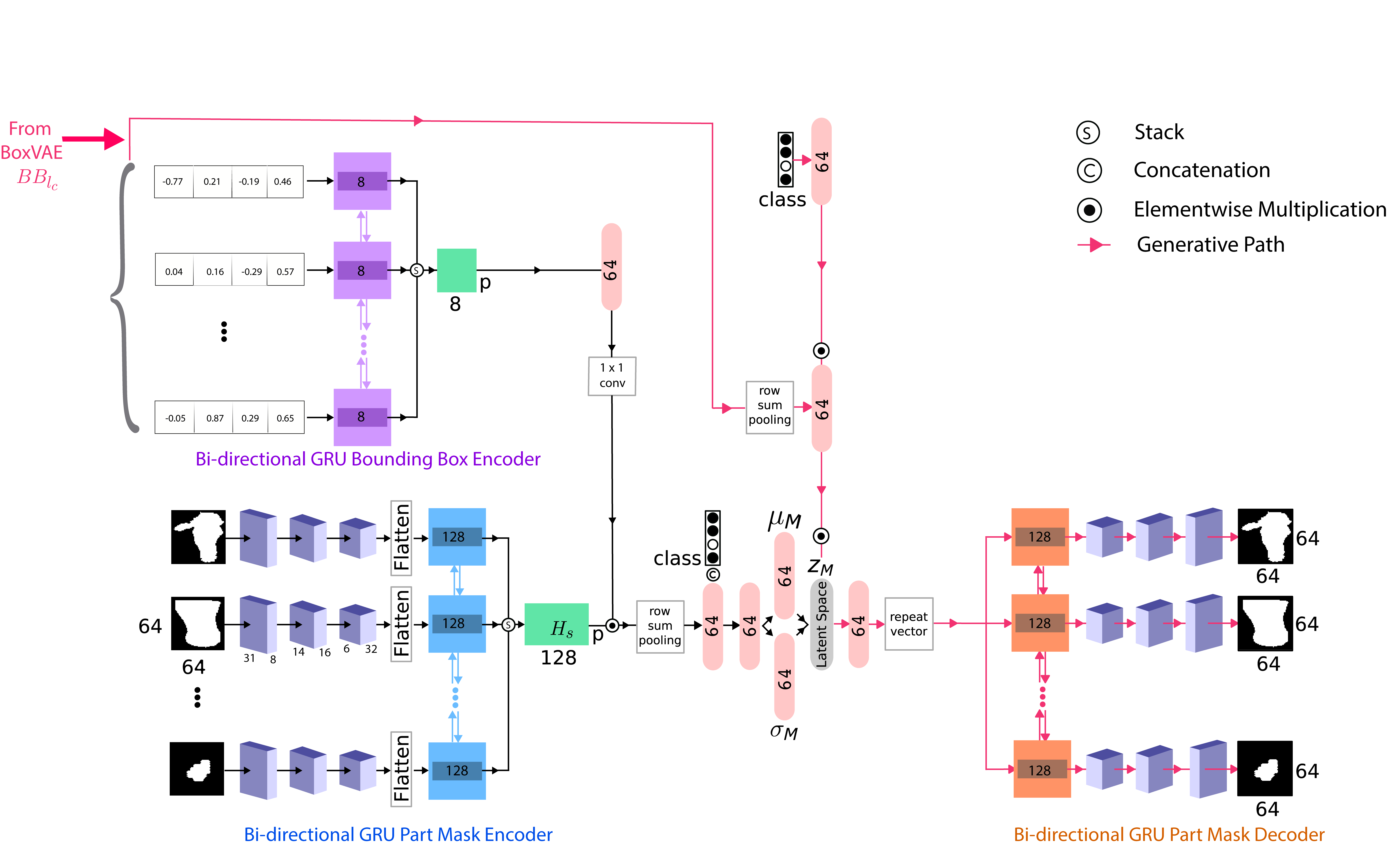}
  \caption{The architecture for LabelMapVAE (Sec. \ref{sec:labelmapvae}) trained to generate per-pixel map for each part conditioned on object class and the part-labeled bounding box layout. After generation, the bounding boxes are used to compose the object in terms of appropriately re-scaled part masks. The pink arrows indicate the data flow for the generative model.}
  \label{fig:labelmapvae}
\end{figure*}

We design LabelMapVAE (Fig. \ref{fig:labelmapvae}) as a conditional-VAE which learns to stochastically generate object layouts, but now as part-label maps. To guide the label map generation in a class-aware and part-aware manner, we use feature representations corresponding to object category $c$ and the bounding box layout $\mathcal{BB}_{l_c}$ generated by BoxVAE (Sec. \ref{sec:boxvae}). Note that $\mathcal{BB}_{l_c}$'s first column indicates the presence or absence of a part in the final layout. Unlike the bounding box which represents a coarse specification of the object, generating the object label map requires spatial detail for each part to be represented accurately. To meet this requirement, we perform encoding and decoding of the label map one part at a time. As before, we first describe the encoder architecture and then describe the decoder.

\noindent \textbf{Encoding the label map:} During encoding, the spatial binary mask for each part is resized to a fixed size ($64 \times 64$). The global consistency of label map is encouraged by two design choices. First, we aggregate the per-part CNN-based feature representations of individual part masks autoregressively using a bi-directional Gated Recurrent Unit (GRU) (color-coded blue in Fig. \ref{fig:labelmapvae}). Second, the hidden-state representations from each unrolled GRU unit are stacked to form a $\mathsf{p} \times h_s$ representation $H_s$ where $h_s=128$.

The feature representations for each part in the bounding boxes $\mathcal{BB}_{l_c}$ generated by BoxVAE are obtained using another bi-directional GRU  (color-coded purple in Fig. \ref{fig:labelmapvae}). An aggregation and stacking scheme similar to the one used for part masks is used here as well to obtain a $\mathsf{p} \times h_b$ feature matrix where $h_b=8$. Using $1 \times 1$ convolutions, the latter is transformed to a $\mathsf{p} \times h_s$ representation $H_b$. To impart bounding-box based conditioning, we use $H_b$ to multiplicatively gate the intermediate label-map feature representation $H_s$. The resulting $\mathsf{p} \times 128$ representation is pooled across rows and gated using category information. The resulting $128$-dimensional feature is ultimately mapped to the same-dimensional parameters of a diagonal Gaussian distribution ($\mu_{M},\sigma_{M}$).

\noindent \textbf{Generating the part label map:} The decoder maps the sampled latent variable $z$ to a conditional data distribution $p_{\theta_m}(M|z,\alpha_{bb},\alpha_{c})$ over the sequence of label maps $M = \{m_1,m_2,\ldots,m_\mathsf{p}$\} with $\theta_m$ representing the parameters of the decoder network. $\alpha_c$ and $\alpha_{bb}$ respectively represent the conditioning induced by feature representations of object category $c$ and the stochastically generated bounding box representation $\mathcal{B}_{l_c}$.
The gated latent vector is mapped to a $128$-dimensional feature $z_g$ which is replicated $\mathsf{p}$ times and fed to the decoder bi-directional GRU (color-coded orange in Fig. \ref{fig:labelmapvae}).
The hidden state of each unrolled GRU unit is subsequently decoded into individual part maps. We model the distribution of part maps as a factored product of conditionals: $p_{\theta_m}(M|z,\alpha_{bb},\alpha_{c}) = \prod_{k=1}^{\mathsf{p}} p(m_k | m_{1:(k-1)}, z, \alpha_{bb}, \alpha_{c})$. In turn, we model each part map as $p(m_k | m_{1:(k-1)}, z, \alpha_{bb}, \alpha_{c}) = \mathsf{ Softmax}(Z_{\theta}^k)$ where $Z_{\theta}^k$ represents the logits obtained from $k$-th part's GRU-CNN decoder. The part maps are then scaled and placed at locations specified by the corresponding bounding boxes in a specific order to obtain the final generated object label map.

\section{Experiments}
\label{sec:experiments}

\subsection{Implementation Details:}
\label{sec:impldetails}

All the components of the architecture (BoxVAE and LabelMapVAE) are trained using the standard approach of maximizing the ELBO as mentioned in Sec. \ref{sec:vae} and with Adam optimizer~\cite{kingma2014adam}. For hyperparameter $\lambda$ which trades off reconstruction loss and KL regularization term, we employ a cyclic annealing schedule~\cite{fu2019cyclical}. This is done to mitigate the possibility of the KL term vanishing and to make use of informative latent representations from previous cycles as warm restarts. In addition, we impose a constraint over the difference of training and validation losses. Whenever the difference increases above a threshold ($0.1$ in our case), the coefficient of KL regularization term is frozen and prevented from increasing according to the default annealing schedule. This condition is maintained until the loss difference comes below the threshold limit. By doing so, we avoid overfitting and mode collapse. BoxVAE is trained with a learning rate of $10^{-4}$ for $300$ epochs using a mini-batch of size $32$. LabelMapVAE is trained with a learning rate of $10^{-3}$ for $110$ epochs using a mini-batch size of $8$.

\noindent Note that the process of generation has already been described before in Sec. \ref{sec:overview}.

\noindent \textbf{Dataset:} To train OPAL-Net, we use the PASCAL-Part dataset~\cite{chen2014detect}, containing $10{,}103$ images across $20$ object categories annotated with part labels at pixel level. We select the following $10$ object categories: \texttt{cow, bird, person, horse, sheep, cat, dog, airplane, bicycle, motorbike}. The individual objects are cropped from the dataset images and centered. To augment images and associated part-label maps, we apply translation, anisotropic scaling for each part independently and also collectively at object level. We also employ horizontal mirroring of objects. The objects are then normalized with respect to the minimum and maximum width across all images such that all objects are centered in a $[-1,1] \times [-1,1]$ bounding box. We use $75\%$ of the images for training, $15\%$ for validation and the remaining $10\%$ for quantitative evaluation.

\section{Baseline Generative Models}
\label{sec:baselines}

\begin{figure*}[!ht]
  \centering
  \includegraphics[width=\textwidth]{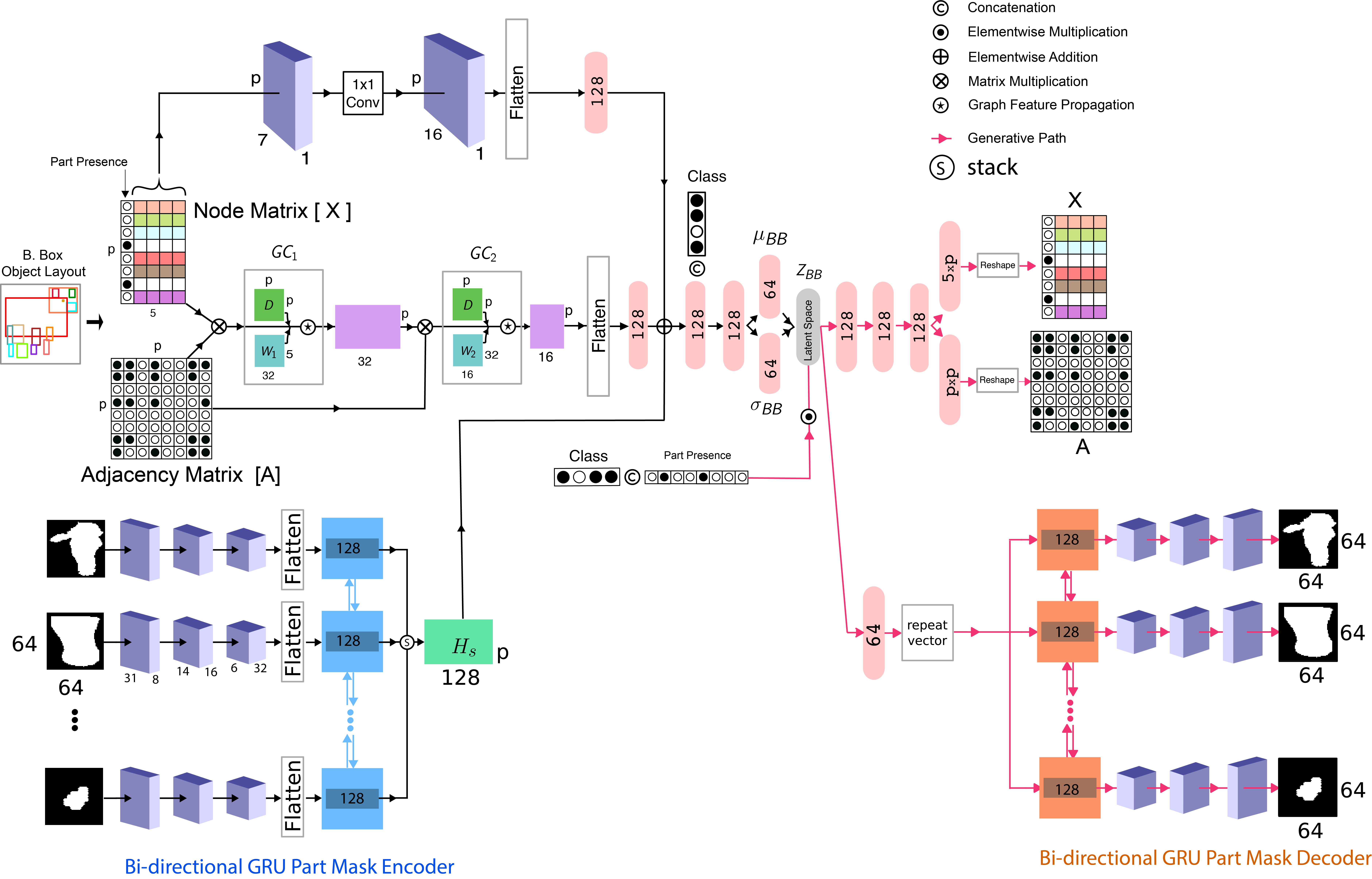}
  \caption{The architecture for Box-Mask-VAE baseline which simultaneously generates bounding boxes and object part masks. The pink arrows indicate the data flow for the generative model.}
  \label{fig:boxmaskvae}
\end{figure*}

\subsection{Box-Mask-VAE (BM-VAE)}

This represents an alternative paradigm to the proposed sequential approach of first generating bounding boxes and generating part masks conditioned on the former. Instead, the bounding boxes and part masks are \textit{simultaneously} encoded using encoder architectures from BoxVAE and LabelMapVAE (Figure~\ref{fig:boxmaskvae}). The generations are enabled within a VAE framework. Paralleling the encoders, bounding boxes and object masks are simultaneously decoded using decoder architectures from BoxVAE and LabelMapVAE.

\noindent Implementation Details: The model uses Adam optimizer with a learning rate of $10^{-3}$ and is trained for $110$ epochs using batch size of $8$.

\subsection{Box-Shape-LSTM (BS-LSTM)}

\begin{figure*}[!ht]
  \centering
  \includegraphics[width=\textwidth]{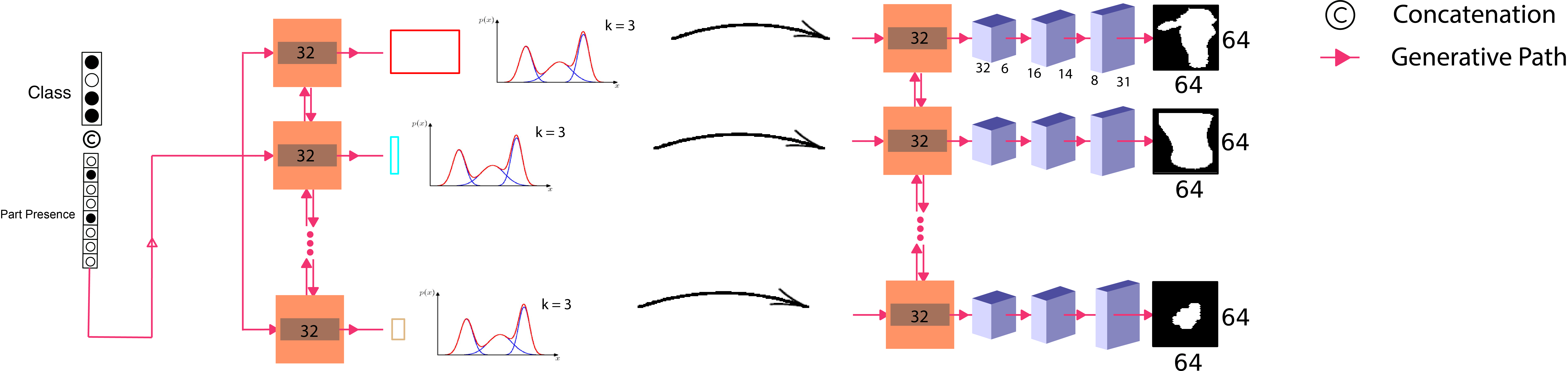}
  \caption{The architecture for Box-Shape-LSTM baseline. The curved black arrow represents sampling from a Gaussian Mixture Model. The Box LSTM is configured to output the parameters of the mixture model for each part. The mixture model itself is defined over the distribution of bounding boxes of parts. The sampled per-part bounding boxes are fed to Shape LSTM which generates corresponding part masks.}
  \label{fig:boxshapevae}
\end{figure*}

This baseline is adapted from Hong et al.~\cite{hong2018inferring}. In their approach, an encoding of a scene description text is fed to a bi-directional LSTM (Box-LSTM) which generates a sequence of bounding boxes and associated object labels. The resulting box sequence is fed to another bi-directional LSTM (Shape-LSTM) which generates a sequence of object masks. We modify the approach by (i) replacing text encoding with an object category and part list encoding (ii) having Box-LSTM, Shape-LSTM generate part bounding boxes, part masks respectively (Figure ~\ref{fig:boxshapevae}).

\noindent Implementation details: Each LSTM cell of the BS-LSTM has a hidden vector of size 32, the model uses  Adam optimizer with a learning rate of $10^{-5}$ and is trained for 300 epochs using batch size of $32$.

\subsection{Conditional Gumbel-GAN (CG-GAN)}

\begin{figure*}[!ht]
  \centering
  \includegraphics[width=\textwidth]{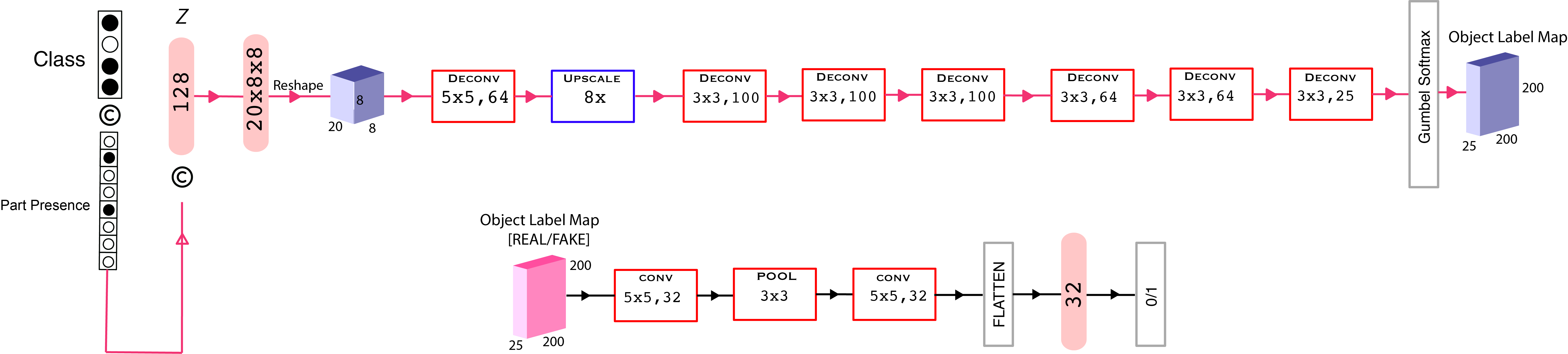}
  \caption{The architecture for CG-GAN baseline. Conditioned on object category and list of parts, the Generator network stochastically generates a feasible object label map. The Discriminator network (towards bottom) attempts to distinguish between generated label maps and those from the training data.}
  \label{fig:cggan}
\end{figure*}

In this conditional GAN-based approach, a random noise vector is used to generate a per-pixel object label map directly. The object category and part-label vector are used as conditioning attributes (Figure~\ref{fig:cggan}). Generating a label map amounts to sampling from a discrete distribution. To achieve this, we use the Gumbel-softmax to model the generator output~\cite{kusner2016gans}.

\begin{figure*}[!ht]
  \centering
  \includegraphics[height=0.9\textheight,width=0.95\textwidth]{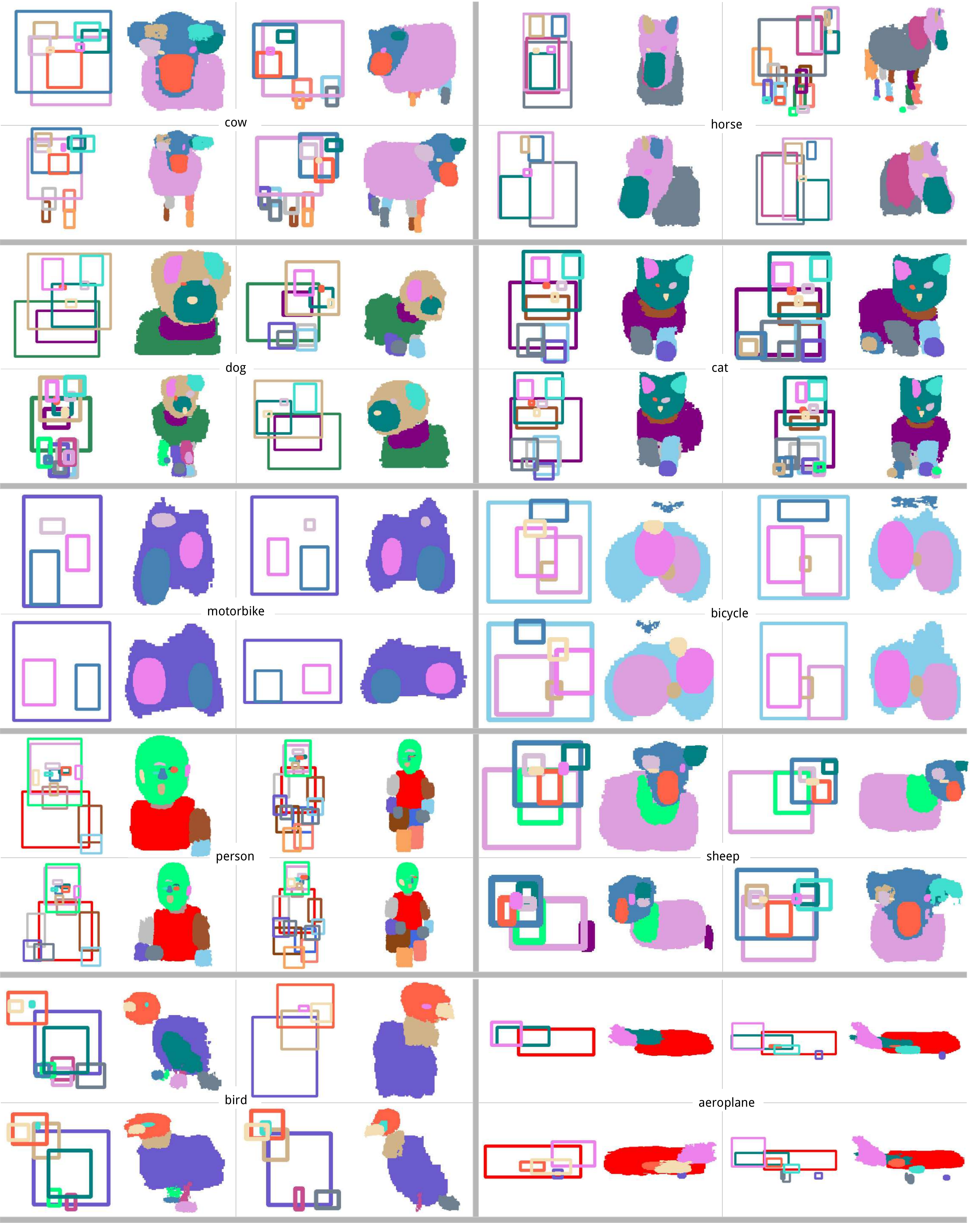}
  \caption{Sample generations from our model (OPAL-Net). Each panel shows the bounding box generated by $\mathsf{BoxVAE}$ and the associated object layout generated by $\mathsf{LabelMapVAE}$. Note that the generations are conditioned by object category and an associated list of parts (not shown to reduce clutter).}
  \label{fig:opalgen}
\end{figure*}

\noindent Implementation Details: The CG-GAN uses a Gumbel-softmax layer to sample from a discrete distribution as the output generated from the model are discrete class masks. Instead of one-hot vector, Gumbel softmax, a soft version of softmax is used. It samples one-hot encoding according to the current learned distribution.

Here we consider a d-dimensional vector $p$ specifying the  probabilities for a multinomial distribution on $y$ with $p_i$ = $p(y_i = 1), i = 1,\ldots d$. We consider a one-hot-encoding
$d$-dimensional vector $y$ and a continuous $d$-dimensional vector $h$, which is used by the softmax function to parameterize a multinomial distribution,i.e. $p = softmax(h)$ and the softmax function being:

\begin{equation}
    softmax(h)_i = \frac{exp(h_i)}{\sum_{j=1}^{K} exp(h_j)}
    \label{eqn:sm}
\end{equation}

\noindent and

\begin{equation}
y = \mathsf{one-hot}(\arg \max_i (h_i + g_i))
\label{eqn:comb}
\end{equation}

Now, sampling $y$ according to the above equation is identical to sampling from the previous multinomial distribution with probability vector given by Equation~\ref{eqn:sm}. Here, the $g_i$ follow a Gumbel distribution and are independent, with zero location and unit scale.

The generated y in Equation~\ref{eqn:comb} has zero gradient zero with respect to $h$ since the one-hot (arg max(·)) operator is not differentiable. This can be approximated by the operator with a differentiable function based on the softmax transformation~\cite{jang2016categorical}. The corresponding probability distribution, parameterized by $\tau$ (the so-called temperature term) and $h$ is called the Gumbel-softmax~\cite{kusner2016gans}.

\begin{equation}
y = softmax(\frac{1}{\tau(h+g)})
\end{equation}

\noindent Implementation details: The GAN is trained on conventional lines. The discriminator consists of convolutional layers which finally predicts if the mask generated is fake or real. Both the generator and discriminator use Adagrad optimizer with a learning rate of $10^{-2}$ and the model is trained for $110$ epochs using batch size of $8$.

\subsection{Results}
\label{sec:results}

For a given generative model, we use the part-presence lists from the test set and generate corresponding samples for each category.

Sample OPAL-Net generations, conditioned on object category and associated part lists, can be viewed in Figure~\ref{fig:opalgen}. The results demonstrate the ability of our unified model to generate diverse, good quality layouts for multiple object categories even when training data is relatively limited.

\begin{figure*}[!ht]
  \centering
  \includegraphics[width=\textwidth]{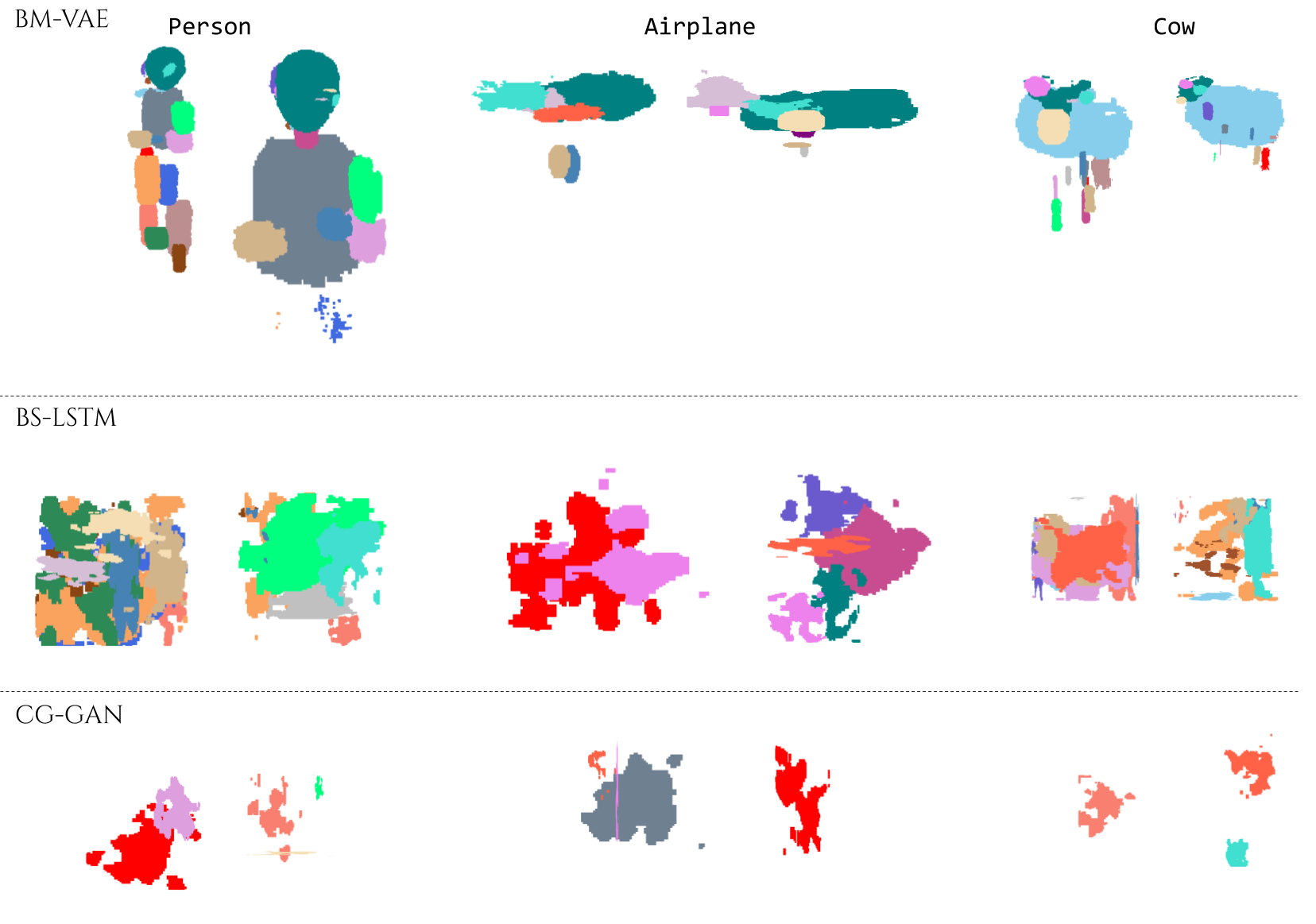}
  \caption{Sample generations for three object categories from the three baselines described in Section~\ref{sec:baselines} -- BM-VAE, BS-LSTM, CG-GAN. The part presence list for each generation is obtained from a randomly chosen sample in the test set.}
  \label{fig:base}
\end{figure*}

Each baselines is competitive and representative of predominant approaches (cGAN, VAE, LSTM) used for generative models. However, they have crucial shortcomings. The GAN-based approach (CG-GAN), which involves one-shot, direct generation of label maps fails since it cannot decouple part geometry and appearance. CG-GAN also fails to reconcile the large range in parts and their relative spatial footprints across multiple categories. Even though BM-VAE baseline contains the same core components present in OPAL-Net, its inferior performance reinforces the importance of decoupling geometry and appearance. BS-LSTM baseline does incorporate decoupling. However, the sequential modelling of part geometry relationships induced by LSTM is not powerful enough to characterize the wide variety of complex layouts that manifest in part-based object representations. This is not an issue in OPAL-Net due, in part, to the GCN-based layout geometry representation. The generations from the CG-GAN and BS-LSTM are practically unusable (see Figure ~\ref{fig:base}), while those from BM-VAE, while somewhat similar to OPAL-Net, render objects as disconnected part groups.

Similarly, the results for ablative variants highlight the importance of key design and optimization choices in OPAL-Net (Section~\ref{sec:OPAL-Net}).

\subsection{Interactive Modification}

\begin{figure*}[!ht]
  \centering
  \includegraphics[width=\textwidth]{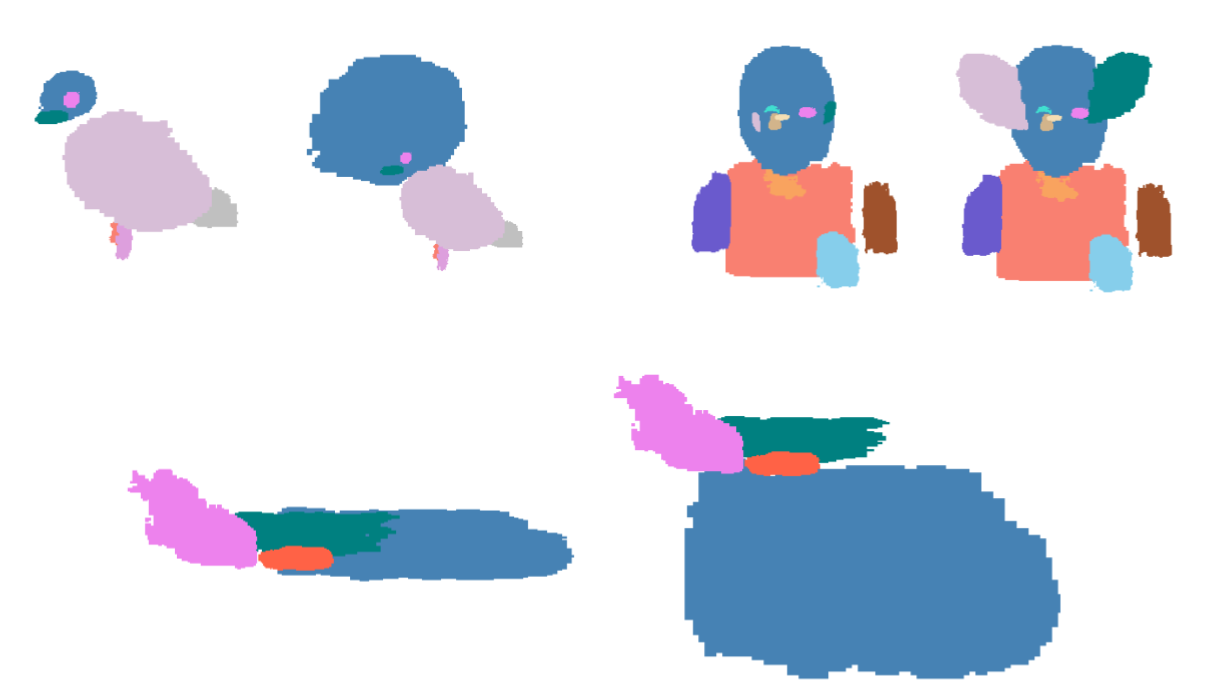}
  \caption{An illustration of interactive modification. For each object, the figure on the left is generated by OPAL-Net unaltered. The figure on the right represents the layout generated when bounding boxes of certain parts in the object are altered in an exaggerated manner. Note the unusually large head of \texttt{bird}, unusually large ears of \texttt{person} and the unusually large fuselage of \texttt{airplane}.}
  \label{fig:caricatures}
\end{figure*}

\noindent \textbf{Manipulating bounding boxes for caricature mask generations:} To showcase the versatility of our framework for interactive modification, we conducted the following experiment: After sampling the bounding box representation from $\mathsf{BoxVAE}$, we modify the position and aspect ratio of part bounding boxes. The resulting boxes are processed as usual by $\mathsf{LabelMapVAE}$ to generate object layouts. This feature is useful in generating cartoon-style, caricature layouts. Some examples can be viewed in Figure~\ref{fig:caricatures}.

\noindent \textit{Adding new parts by label:} Consider an object $d$ from the test set. We obtain a sample $z_d$ from $\mathsf{BoxVAE}$'s encoder distribution $q_{\phi}(z|\mathsf{d})$. A part originally not present in the test object, specified by the user, is included in the part-presence list. The new list and test object's category is used to condition $z_d$. The `conditoned' version of $z_d$ is then used to obtain a bounding box representation via $\mathsf{BoxVAE}$'s decoder. Only the bounding box for the newly added part is added afresh while the rest of the bounding boxes are used as is from the test object.

\begin{figure*}[!ht]
  \centering
  \includegraphics[width=\textwidth]{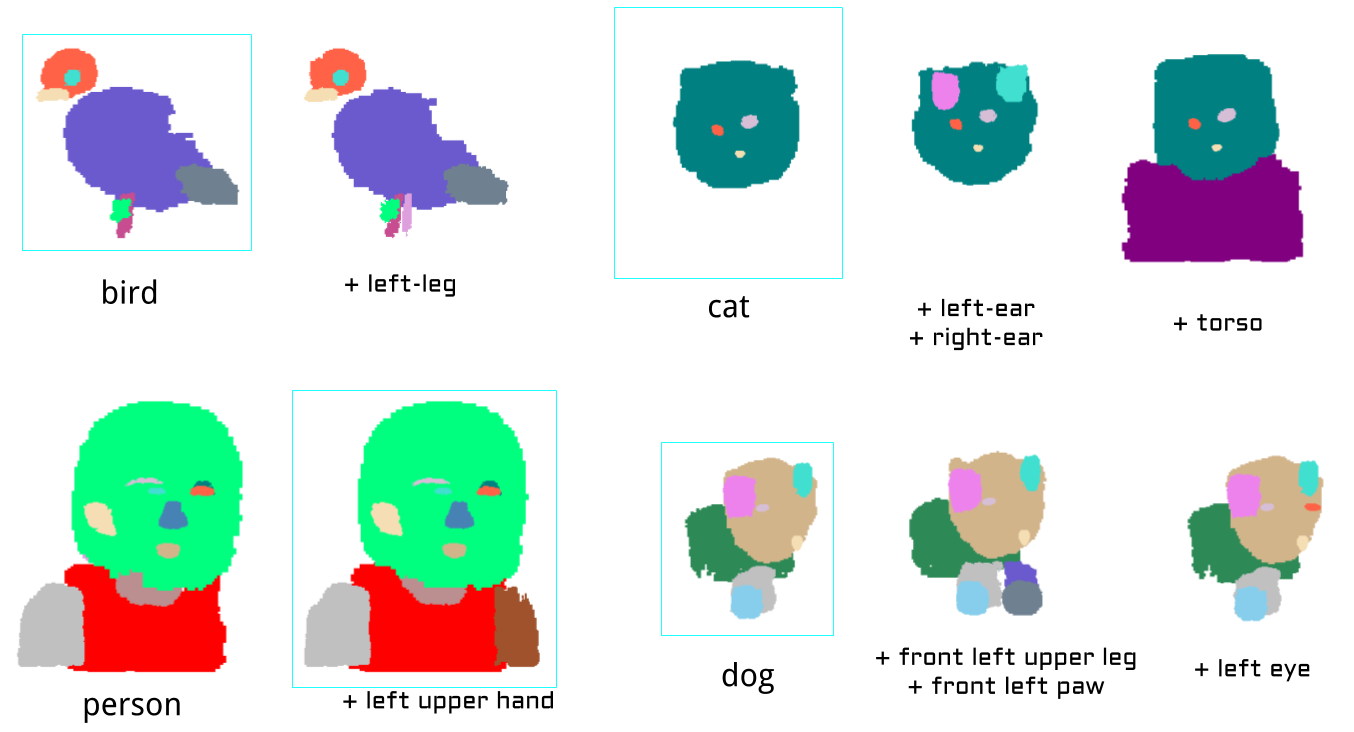}
  \caption{For each test layout image (cyan border), a new part label is specified for inclusion via $\mathsf{BoxVAE}$'s conditioning. This causes $\mathsf{BoxVAE}$ to `hallucinate' the newly added part. The usual box-conditioned generation of masks in $\mathsf{LabelMapVAE}$ results in a plausible part map of the user-specified part added to the test image's layout. The added parts are mentioned with prefix $+$.}
  \label{fig:addparts}
\end{figure*}

\begin{figure*}[!ht]
  \centering
  \includegraphics[width=\textwidth]{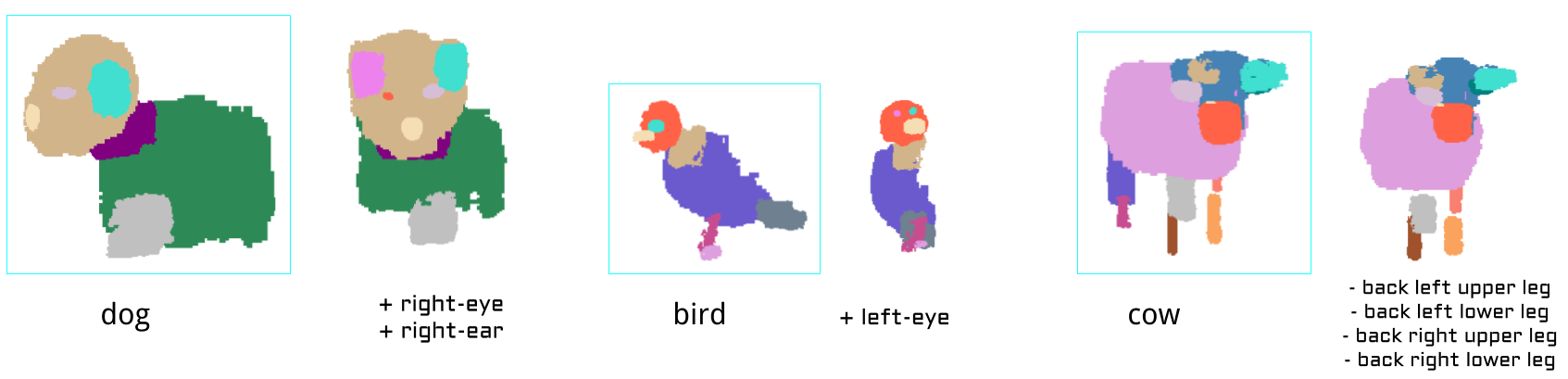}
  \caption{The depictions here share the same description as Figure~\ref{fig:addparts}. The distinctive feature is that addition or deletion of certain parts sometimes results in part lists which are highly correlated with specific camera viewpoints.}
  \label{fig:addparts2}
\end{figure*}

A similar procedure, but with the new list and the new-part-added bounding box representation is used by $\mathsf{LabelMapVAE}$ to generate the mask for added part. As with bounding boxes, this mask is added afresh while retaining the rest of the original part masks from the test object. The results can be viewed in Figure~\ref{fig:addparts}. Sometimes, addition or deletion of certain parts sometimes results in part lists which are highly correlated with specific camera viewpoints. The net effect resembles a novel rendering of original object with a viewpoint correlated with list of final parts. Examples can be seen in Figure~\ref{fig:addparts2}.

\section{Conclusion}

In this paper, we have presented a hierarchical generative framework for objects. Our model, OPAL-Net, generates diverse part-based object layouts from multiple categories using a single unified architecture. The strict constraints between object parts, variety in layouts and extreme part articulations generally make multi-category object generation a very challenging problem. Through our design choices involving GCNs, VAEs and guidance using object attribute-based conditioning, we show that this problem can be tackled using a single unified model. An added advantage of our hierarchical model is that it enables efficient processing and scaling with inclusion of additional object categories in future.

\clearpage

\bibliographystyle{splncs}
\bibliography{egbib}

\end{document}